\documentclass[12pt,twocolumn]{article}
\usepackage{geometry}

\usepackage{array}
\usepackage[table]{xcolor}

\usepackage[T1]{fontenc}
\usepackage[utf8]{inputenc} 
\usepackage{float}
\usepackage[justification=justified,singlelinecheck=false,font=small,labelfont=bf,textfont=it]{caption}
\usepackage{enumitem}
\usepackage{pifont}

\usepackage[Algorithm,ruled]{algorithm}
\usepackage{algorithmicx}
 \usepackage{algpseudocode}
 
\usepackage{boxedminipage}
\usepackage[american]{babel}
\usepackage{hyperref}
\usepackage{graphicx}
\usepackage{subcaption}
\usepackage{color}
\usepackage{amsmath}
\usepackage{mathptmx}
\usepackage{bm}
\usepackage{caption}

 \usepackage[usenames,dvipsnames]{pstricks}
 \usepackage{epsfig}
\usepackage{pst-grad} % For gradients
\usepackage{pst-plot} % For axes

\title{\textbf{Interactive multiclass segmentation using superpixel classification} \\[1ex] \large A new fast and efficient multiclass interactive segmentation method}
\author{\small \textbf{Bérengère Mathieu} \\ \small Université de Toulouse\\ \small UPS ;  IRIT \\ \small 118 route de Narbonne,\\ \small F-31062 Toulouse, France \\ \small  \emph{berengere.mathieu@irit.fr} \and \small  \textbf{Alain Crouzil} \\ \small  Université de Toulouse\\ \small UPS ;  IRIT \\ \small 118 route de Narbonne,\\ \small F-31062 Toulouse, France \\ \small  \emph{alain.crouzil@irit.fr} \and \small  \textbf{Jean-Baptiste~Puel} \\ \small Université de Toulouse\\ \small ENFA ;  IRIT \\ \small 118 route de Narbonne,\\ \small F-31062 Toulouse, France \\ \small  \emph{jean-baptiste.puel@irit.fr} }	

\definecolor{myblue1}{rgb}{0.68, 0.79, 0.86}
\definecolor{myblue2}{rgb}{0.58, 0.69, 0.86}
\definecolor{nicegray}{rgb}{0.8,0.8,0.8}

\begin{document}
\maketitle
\begin{abstract}
\textit{
This paper adresses the problem of interactive multiclass segmentation. We propose a fast and efficient new interactive segmentation method called Superpixel Classifica\-tion-based Interactive Segmentation (SCIS). From a few stro\-kes drawn by a human user over an image, this method extracts relevant semantic objects. To get a fast calculation and an accurate segmentation, SCIS uses superpixel over-segmentation and support vector machine classification. In this paper, we demonstrate that SCIS significantly outperfoms competing algorithms by evaluating its performances on the reference benchmarks of McGuinness and Santner.}

\end{abstract}

\section{Introduction}
\begin{figure}[!t]
	 \begin{subfigure}[t]{0.45\textwidth}	
	 \begin{center}
		\scalebox{0.4}{
			\includegraphics{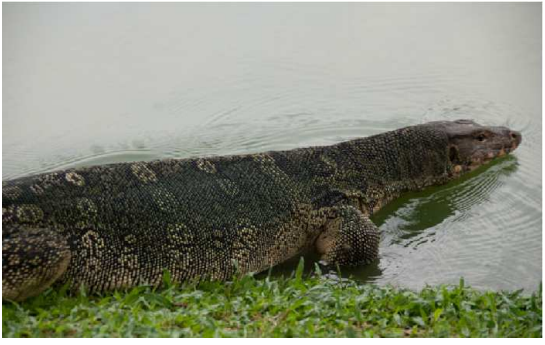}
		 }	 
	 \end{center}
		 \caption{Image of a reptile in the water, from Santner benchmark \cite{santner2010Interactive}.}
		 \label{fig:divSef00}
	\end{subfigure}	
\\[1ex]
	\begin{subfigure}[t]{0.45\textwidth}	
	 \begin{center}
		\scalebox{0.4}{
			\includegraphics{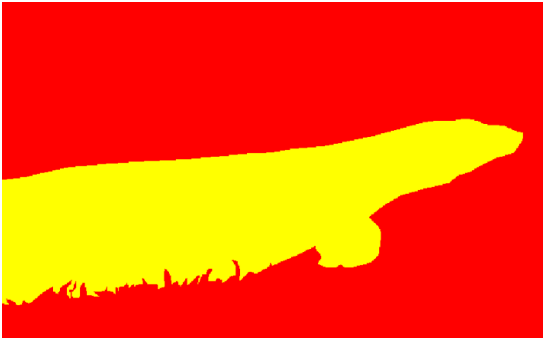}
		}	
	 \end{center}
        \caption{Segmentation for image editing application: the main element (the reptile)  is extracted from the background.}
        \label{fig:divSef01}

	\end{subfigure}
\\[1ex]
	 \begin{subfigure}[t]{0.45\textwidth}	
	 \begin{center}
		\scalebox{0.4}{
			\includegraphics{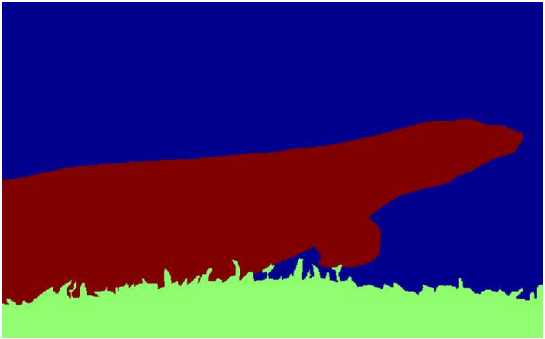}
		 }
	 \end{center}
		\caption{Segmentation for image understanding application: each region corresponding to a semantic element (reptile, water and grass) is extracted.}
		\label{fig:divSef02}
	\end{subfigure}
      \caption{Diversity of possible segmentation results.}
      \label{diversityOfSegmentationResults}
\end{figure}

Image segmentation is still a challenging research in the image analysis community. Its goal is to group similar and neighboring pixels in order to partition the image into structures corresponding to coherent elements. However, depending on the application area, for a same image the term  ``coherent elements'' can have different meanings. An example is given in Figure  \ref{diversityOfSegmentationResults} which shows for the same image two possible segmentations corresponding to two different applications: image editing and image understanding. In the first case, the segmentation algorithm is supposed to find only two regions: the main object and the background. In the second case the number of regions is not \textit{a priori} known: the algorithm is supposed to find all relevant semantic objects.

This example illustrates the complexity of finding a segmentation algorithm able to deal with the variety of contexts of use.  Interactive segmentation bypasses this difficulty  by asking some additional information to the user. Unfortunately, in spite of their potential great flexibility, the majority of interactive segmentation methods are designed to only solve binary classification problems \cite{adams1994seeded,blake2004interactive,boykov2001interactive,chapelle2004semi,grady2006random,mcguinness2010comparative,zhou2012interactive}. In this article, we introduce the Superpixel Classification-based Interactive Segmentation (SCIS) algorithm, an efficient, simple and fast interactive segmentation method able to find several elements in a image.

In this paper, we begin with an overview of the interactive segmentation field focused on user interaction, visual features and segmentation problem formulation. We then describe SCIS, the proposed multiclass interactive segmentation method, first its generic algorithm and then some possible variations that we later evaluate over a set of data. We conclude by evaluating the performance of the most efficient variant (SCIS-SVM) by comparing it to four other interactive methods previously described. We conclude with the promising results of SCIS, its weaknesses and some ideas to compensate them.

\section{Related works}
\label{sec:previousWork}

As its name suggests, interactive segmentation is a semi-automatic segmentation method. The user chooses some pixels (named seeds) and indicates for each of them the element to which it belongs.  Features (location, color, texture, etc.) of desired regions are deduced by analyzing these seeds. Usually, adding or removing some seeds can improve the produced result, allowing the user to get any desired segmentation results. 

Understanding previous works in interactive segmentation involves answering the three following questions: how does the user interact with the method? Which features are used to describe regions? How is the segmentation problem modeled and solved?

\subsection{User interaction}
 It is generally accepted that segmentation methods can be classified into edge-based algorithms and region-based algorithms. 

When an interactive segmentation method searches for boundaries of the different elements of a picture, the user must usually give some points of these boundaries. One of the most representative edge-based interactive segmentation method is Intelligent Scissors (Magnetic Lasso) \cite{mortensen1998interactive}. However, for highly textured images, a lot of boundary points are often required,  making user interaction tedious.

Regarding most of the region-based methods, user must draw some strokes on the different regions \cite{boykov2001interactive,santner2009interactive,xu2008transductive,zhou2012interactive}. With this modality of interaction, if picture elements are small when compared to the brush size, user can unintentionally give wrong seeds. Some methods attempt to deal with this noise \cite{xu2008interactive,xu2008transductive} but, despite everything, if the application provides a zoom and different brush radii, results can be significantly better.

\subsection{Visual features}

The  most frequently used features to describe regions are pixels locations, gray levels and colors in RGB (Red Green Blue), HSV (Hue Saturation Value) or Lab color spaces. Some authors, such as Boykov \textit{et al.} \cite{boykov2001interactive} or Grady \cite{grady2006random}, simply use pixel gray levels and locations to segment images in several homogeneous regions. Other methods \cite{blake2004interactive,duchenne2008segmentation,friedland2005siox,kuang2012learning,li2004lazy,rother2004grabcut} use pixel colors. For example, Blake \textit{et al.} \cite{blake2004interactive} characterize each class color distribution as a mixture of randomly generated gaussians. However, defining a color model for each class is not necessary: Duchenne \textit{et al.} \cite{duchenne2008segmentation} directly use the image colors in a fixed-size window surrounding each pixel.

Lately, some works \cite{kuang2012learning,santner2010Interactive,zhou2012interactive}  have started to investigate the use of texture features with descriptors like textons \cite{malik1999textons}, structure tensors \cite{rousson2003active} or histograms of oriented gradients \cite{dalal2005histograms}.

\subsection{Formulation of the segmentation problem}
 According to the way they choose to model and solve the segmentation problem, existing methods can be classified into two categories: region growing algorithms \cite{ning2010interactive,salembier2000binary,zhou2012interactive} and energy minimization algorithms \cite{boykov2001interactive,duchenne2008segmentation,grady2006random,santner2010Interactive}. We will now focus on each of these categories and describe some relevant algorithms. 

\subsubsection{Region growing algorithms}
In region growing algorithms, seeds are the starting point of searched regions. These embryonic regions are then completed by clustering neighboring pixels which satisfy a similarity criterion. Then, features of the new regions are updated. This process is iterated until all the pixels belong to a region. Algorithms differ by the choice of the merging order, the merging criterion and the way by which region features are updated when a new pixel is added.

\paragraph{Segmentation using Binary Partition Trees (BPT):}The idea to use a binary partition tree algorithm to segment an image is first proposed by Salembier \textit{et al.} \cite{salembier2000binary} and then expanded by Adamek in \cite{adamek2006using} and McGuinness \textit{et al.} in their review \cite{mcguinness2010comparative}. 
\begin{sloppypar}
McGuinness \textit{et al.} use the RSST (Recursive Shortest Spanning Tree) algorithm \cite{morris1986graph} to create the binary partition tree. Then, they transform this tree into a binarization, using the algorithm suggested by Salembier \cite{salembier2000binary}. First, acccording to the seeds given by a user, some leaf nodes of the binary tree are labeled. Then these labels are propagated to parent node, until a conflict occurs when the two children of a node have different labels. In this situation, the parent nodes is marked as conflicting. Finally, all non-conflicting nodes propagate their label to their children.
\end{sloppypar}

At the end of this stage, some sub-trees can remain unlabeled. McGuinness \textit{et al.} follow the approach proposed in \cite{adamek2006using} by labeling each unclassified region with the label of an adjacent previously classified region. When several adjacent regions with different labels are candidates to label the unclassified region, the closest one according to the Euclidean distance is chosen.

BPT method obtained the best scores in  the McGuinness \textit{et al.} review on interactive segmentation methods to solve binarization problems.

\paragraph{Contour Detection and Hierarchical Image Segmentation (CDHIS):}
~
\\
Interactive segmentation method of Arbelaez \textit{et al.} \cite{arbelaez2011contour} works in three stages: first, boundaries are extracted, then a hierarchical segmentation is computed and finally this segmentation tree provides the starting point for a user-assisted segmentation.  

Arbelaez \textit{et al.} use both local and global information to search boundaries in the image. Local information is the probability for a pixel to belong to a contour according to color and texture of its neighborhood. In practical terms, giving a disc centered on the pixel $p$ and split by a diameter at angle $\theta$,  Arbelaez \textit{et al.} analyze the  dissimilarity  between the histograms ($g$ and $h$) of the two half-disc using $\chi^{2}$ distance:
\begin{equation}
\chi^{2}(g,h) = \frac{1}{2}\sum_{i} \frac{(g(i) - h(i))^{2}}{g(i) + h(i)}.
\end{equation}

For each diameter orientation, this dissimilarity is computed for four different features (the three Lab channels and a  texton identifier associated to each pixel) and for three different scales (changing the disc radius). So,  each pixel, each orientation,  scale and  channel give a local signal $G_{i,\sigma(i,s)(x,y,\theta)} $ where  $i$ is the feature channel index,  $s$ is the scale index and $\sigma(i,s)$ is the disc radius.

For each pixel at location $(x,y)$, Arbelaez \textit{et al.} compute:
\begin{equation}
mPb(x,y)=\sum_{s} \sum_{i} \alpha_{i} G_{i,\sigma(i,s)(x,y,\theta^{*})} 
\end{equation}
with $\theta^{*}$ the orientation giving the highest signal and $\alpha_{i}$ the weights of each local signal contribution. Then, they search  for the eigenvectors $v_{0},...,v_{n}$ corresponding to the $n+1$ smallest eigenvalues of the system $(D-W)v = \lambda D v$ with:
\begin{itemize}
\item $W$ a sparse symmetric matrix such as \\ $W_{i,j} = exp ( - \frac{ 1 } {\rho} \underset{p \in \overline{ij}} \max \lbrace mPb(p) \rbrace ) $;
\item $ \overline{ij}$ the line segment connecting pixels $i$ and $j$;
\item $D$ a diagonal matrix such as  $D_{i,i} = \displaystyle \sum_{j=0}^{N} W_{i,j}$.
\end{itemize}
Arbelaez \textit{et al.} show that these eigenvectors carry contour information: by convolving them with Gaussian directional derivative filters at multiple orientations and by linearly combining the results, they obtain the “spectral” component of their contour detector, denoded by $sPb(x,y,\theta)$.
The final glo\-bal probability for a pixel at coordinates $(x,y)$ to belong to a contour of orientation $\theta$ is given by computing a  sum of local and spectral  signals : $gPb(x,y,\theta)$.

To create the segmentation tree, Arbelaez \textit{et al.} use the maximal response of their contour detector over orientations:
\begin{equation}
E(x,y)=\underset{\theta} \max \lbrace gPb(x,y,\theta) \rbrace.
\end{equation}
The local minima of $E(x,y)$ are taken as seed locations for homogeneous segments and an over-segmentation is constructed using watershed transform on the topographic surface $E(x,y)$. Then, watershed arcs are approximated by line segments, giving for each pixel belonging to the arc an orientation $\theta$. Using this orientation, Arbelaez \textit{et al.} get, for each pixel, a boundary strength $gPb(x,y,\theta)$ and compute for each original arc its weight as the average boundary strength of the pixels it contains. The final hierarchical segmentation is produced by merging regions of the over-segmentation, according to their similarity, which is the average weight of their common boundary. 

Arbelaez \textit{et al.} obtain a segmentation tree. In their interactive segmentation method, seeds given by the user provide a label for some regions of this tree. To complete the segmentation, they give to each unlabeled region the label of the first labeled region merged with it. 

Arbelaez \textit{et al.} did not quantitatively evaluate their interactive segmentation method. They just give the  obtained segmentations for some Berkeley database images\footnote{\url{www.eecs.berkeley.edu/Research/Projects/CS/vision}}. These results seem very promising, nonetheless searching boundaries and computing hierarchical segmentation is a time-consuming task (a few minutes on a desktop computer featuring 2.6 GHz Intel Core i7 processor).

\subsubsection{Energy minimization algorithms }

From the point of view of energy minimization algorithms, the segmentation problem can be represented by a function whose value depends on the image features and a given segmentation. This function achieves its minimum for an optimal segmentation. Seeds given by a user are talen into account as constraints, so a classical way to define a minimization function is to ensure that its value decreases when regions of the segmentation are homogeneous, when regions boundaries are regular and when regions are coherent with seeds. 

\paragraph{Interactive Graph Cuts (IGC):}In \cite{boykov2001interactive} Boykov and Jolly suggest an interactive segmentation method for binary classification problems. They formulate the interactive segmentation problem as the minimization of an energy function grouping hard constraints (seeds provided by user interaction) and soft constraints (similarities between pixels in spatial and intensity domains).  

The cost function to minimize is the following:
\begin{equation}
E(A,I) =\lambda R(A,I) + B(A,I)
\end{equation}
where $I$ is the image,  $A$ is a possible image segmentation,  $R$ is the regional term which evaluates if pixels within a region are coherent with seeds assigned to the region by the user, $B$ is a function which evaluates the boundary relevance.

To find the segmentation $A^{*}$ which minimizes $E$, Boykov and Jolly use a fast min-cut/maxflow algorithm on the graph  $G<V,E>$  such as:
\begin{itemize}
\item $V = V_{p} \cup V_{term}  $, where:
\begin{itemize}
\item $V_{p}= (v_{1}, ..., v_{n})$  is a set of nodes such that each pixel $p_{i}$ is linked to a node $v_{i}$;
\item $ V_{term}= (T,S)$ is a set grouping two  particular nodes: $T$ the sink and $S$ the source; 
\end{itemize}
\item $E = E_{n} \cup E_{t}$, where: 
\begin{itemize}
\item $E_{n} $ is the set of  edges linking $V_{p}$ nodes, under a standard 4- or 8- neighborhood system;
\item $E_{t}$ is the set of edges linking each node in $V_{p}$ to each node in $V_{term}$.
\end{itemize}
\end{itemize}

Each edge is weighted such that $E_{n} $ edges have a strong weight if the two linked nodes correspond to similar pixels and  $E_{t} $ edges have a high weight if the $V_{p}$ node has a strong probability to belong to the foreground when the second node is the source and to the background when the second node is the sink. 

Boykov \textit{et al.} method obtains very good results in the McGuinness interactive segmentation review \cite{mcguinness2010comparative}, but it only solves binary classification problem.

\paragraph{ Texture Segmentation using Random Forests and Total Variation (TSRFTV) :} In \cite{santner2010Interactive}, Santner \textit{et al.} suggest a new interactive multi-class segmentation method in two steps. First, pixel likelihood to belong to each class is computed thanks to  a random forest classifier. Then, an optimal segmentation is found by minimizing an energy which linearly combines a regularization term and a data term:
\begin{equation}
E=\frac{1}{2} \sum_{i=1}^{N} Per_{D}(E_{i};\Omega) + \lambda  \sum_{i=1}^{N}  \int_{E_{i}} f_{i}(x)dx
\end{equation}
where $\Omega$ is the image domain, $E_{i}$ are the $N$ pairwise disjoint sets partitioning $\Omega$ , $Per_{D}$ is a function penalizing the length of the partition and $f_{i}$ is the output of the random forest classifier. 

During the first stage, color and texture of pixels are extracted. Santner \textit{et al.} compared several features and showed that the combination of Lab color space and LBP texture descriptor give the best results. Next the classifier is trained with the seeds. The trained classifier gives for each pixel the likelihood to belong to each class. These likelihoods are used during the second step, into the data term. The regularization term penalizes the boundary length, avoiding noisy segmentation. Santner \textit{et al.} formulate the segmentation problem like a Potts model and solve it using a first order primal-dual algorithm.

According to Santner \textit{et al.} evaluation \cite{santner2010Interactive}, their interactive segmentation method is able to segment a wide variety of images into several classes. However, texture descriptor computation and energy minimization are both time consuming. The process takes more than one second, even for small images ($625 \times 391$ pixels) on a desktop PC. 

\subsection{Remaining issues}

Because it requires iterative manual inputs, an acceptable interactive segmentation method has to minimize the human effort and produce quickly a result. That is to say, the seeds required to obtain a desired result must be neither too accurately localized nor too numerous, and the time to compute a segmentation must aim to a few seconds. However, to the best of our knowledge, a lot of current interactive segmentation algorithms are often time and memory consuming, even if they deal with small images, like images of the Berkeley database\footnote{\url{www.eecs.berkeley.edu/Research/Projects/CS/vision}}. Furthermore, the huge majority of existing interactive segmentation methods deal only with binary classification problems  \cite{adams1994seeded,blake2004interactive,boykov2001interactive,duchenne2008segmentation,friedland2005siox,grady2006random,kuang2012learning,li2004lazy,rother2004grabcut,xu2008interactive}.

In the next section, we will show that it is possible to design an efficient multiclass interactive segmentation algorithm.

\section{Superpixel Classification-based Interactive Segmentation (SCIS)}
\label{sec:SCIS}
In this section, we propose a new region-based segmentation method to partition an image into several regions thanks to the aid of a human operator. 
\subsection{User interaction}
We implemented all the proposed variants of SCIS as Gimp\footnote{\url{www.gimp.org/}} plugins. Seeds are provided thanks to a brush, allowing the user to draw strokes on semantic elements he wants to separate. Seeds for different semantic elements have different colors (for example blue for the sky, green for the grass, pink for a plane, etc.) but two distinct regions can belong to the same class and have seeds with the same color, even if their pixels are not connected. 

\subsection{Algorithm overview}
We consider an image of $M$ pixels  as a set  $P$ of $M$ vectors $P=\lbrace p_{1},\cdots,p_{M} \rbrace$, with $p_{i}$ the vector containing the features of the $i^{\text{th}}$ pixel.  First, we group similar neighboring pixels in small homogeneous sets, named superpixels. For each superpixel we compute its features. 

The second part of our algorithm is iterative: as long as the user adds or removes seeds, the segmentation is updated. We start by analyzing seeds provided by the user to deduce $K$, the number of classes. We create a set of $K+1$ labels $ \lbrace 0, 1,\cdots,K\rbrace$ with $1,\cdots,K$ the labels representing each class and $0$ a void label. So we can assign to each pixel $p$ a label $ j$ such as:
\begin{itemize}
\item $j > 0 $ if $p$ is a seed; 
\item $j = 0$ otherwise.
\end{itemize} 
Then, for each class, we create $S_{j}$, the set of superpixels including at least one pixel with label $j$ and the others with label $j$ or $0$.
Next, we use a learning method for multiclass classification, using, for each class $j$, the features of the superpixel set $ S_{j}$.

At the end of this step, several superpixels are not associated  to a class, because they contain no seed  or because they contain seeds of different classes. We classify these remaining superpixels using the classifier and we assign each pixel to the class of its superpixel. A summary of the proposed method is given in Algorithm \ref{alg:scis}.

\begin{algorithm}[t]
\floatname{algorithm}{\small Algorithm}
\caption{\small Superpixel Classification-based Interactive Segmentation}
\begin{algorithmic}
\State \textsc{Inputs:}
\State Image /* a color image */ 
\State Seeds /* locations and labels of some pixels called seeds */
\State \textsc{Procedure:}
\State Over-segment the image into superpixels and create $S$, the set of all superpixels
\State Extract features of each superpixel
\While{the user gives new seeds}
\State /* Analyze seeds to deduce features of the classes that the user wants to extract */
\State Deduce $K$ the number of classes by analyzing seeds
\For{ each class $j$}
\State Create the set $S_{j}$ of superpixels including only seeds of class $j$ and
\State unlabeled pixels
\EndFor
\State Train the classifier with  $S_{train} =  \underset{j=1,\cdots,K}{ \bigcup}S_{j} $
\State /* Segment the image*/
\For{each superpixel $s \in S - S_{train}$ }
\State Use the classifier to predict the class of $s$ 
\EndFor
\For{each superpixel $s \in S $ }
\For{each pixel $p \in s $}
\State Assign the class of $s$ to  $p$
\EndFor
\EndFor
\EndWhile
\end{algorithmic}
\label{alg:scis}
\end{algorithm}

We will now focus on essential parts of this algorithm, namely the extraction, the description and the classification of the superpixels.

\subsection{Superpixel over-segmentation issues}

As explained above, superpixels are small consistent regions. Superpixels are the result of an over-segmentation of the image: boundaries of elements in the image should match superpixel boundaries, but a same element can be partitioned into several superpixels. Because superpixels are considerably less numerous than pixels, training and using a classifier with superpixels is significantly faster than using pixels. Nevertheless, over-segmentation errors (\textit{i.e.} one superpixel overlapping elements that the user wants to separate) cannot be corrected, no matter the seeds added or removed by the user. 

In the context of interactive segmentation, a good superpixel over-segmentation algorithm must have two opposite qualities: to produce as few as possible superpixels (to reduce the number of elements to classify and the execution time) and to make as few as possible over-segmentation errors to reduce errors in the final segmentation. Unfortunately, reducing the number of superpixels requires to increase their size and by grouping more and more pixels into a same superpixel, the probability of overlapping erros increases dramatically. Moreover, in order to be utilizable by an interactive method, the superpixel extraction has to be fast. 

We compared four superpixel extraction algorithms:  \emph{Fel\-zenzswalb et al.}  algorithm \cite{felzenszwalb2004efficient}, \emph{mean shift} segmentation method \cite{comaniciu2002mean}, \emph{SLIC} method \cite{achanta2010slic} and \emph{Veksler et al.} algorithm \cite{veksler2010superpixels}. Other algorithms have been ruled out according to the Achanta \textit{et  al.} study \cite{achanta2010slic}, showing they are too slow or not accurate enought. We  used ground truth provided by McGuinness \textit{et al.} \cite{mcguinness2010comparative}, which is more accurate than Berkeley ground truth used by Achanta \textit{et al.} For each algorithm, we measured the time required to produce the over-segmentation, the number of superpixels and the rate of incorrectly classified pixels (in a superpixel with a majority of pixels belonging to a given class, the number of pixels belonging to other classes). For each algorithm we tuned the parameters to obtain a compromise between accuracy and speed. For \emph{Felzenzswalb et al.} algorithm, we specified a constant for the threshold function $k=24$ and a minimal size for superpixels of 20 pixels. For \emph{mean shift}, we used a spatial window radius of 3 pixels, a color window radius of 3 pixels and  a minimal size for superpixels of 20 pixels. For \emph{SLIC}, we chose a compactness parameter equal to 10 and an average  superpixel size of 100 pixels. For \emph{Veksler} algorithm, we used a patch size of 20 pixels. 
 Table \ref{tab:resSp} shows the  results of our experimentations.

\begin{table}
	\caption{Comparison of four superpixel over-segmentation algorithms. \emph{Sp} is the average number of superpixels by image, \emph{Time} is the average execution time, in seconds and by image. \emph{Error} is the average percentage of incorrectly classified pixels by image.}
	\label{tab:resSp}
	\begin{center}
	\scalebox{0.5} % Change this value to rescale the drawing.
	{
	\begin{pspicture}(0,0)(16,6)
		\newrgbcolor{myblue1}{0.68, 0.79, 0.86}
		\newrgbcolor{myblue2}{0.58, 0.69, 0.86}
		%premiere ligne
		\psframe[linewidth=0.04,dimen=outer,fillstyle=solid,fillcolor=lightgray](0,6)(16,4)		
		\rput[Bl](2,4.8){\LARGE \textbf{Method}}
		\rput[Bl](8,4.8){\LARGE \textbf{Sp}}	
		\rput[Bl](11,4.8){\LARGE \textbf{Time}}
		\rput[Bl](14,4.8){\LARGE \textbf{Error}}								
		%resultats Felzenzswalb
		\psframe[linewidth=0.04,dimen=outer,fillstyle=solid,fillcolor=myblue1](0,4.04)(16,3.04)
		\rput[Bl](2,3.3){\LARGE \textbf{Felzenzswalb}}
		\rput[Bl](8,3.3){\LARGE 1926}	
		\rput[Bl](11,3.3){\LARGE \textbf{0.2 s}}
		\rput[Bl](14,3.3){\LARGE \textbf{0.5 \%}}	
		%resultat meanshift
		\psframe[linewidth=0.04,dimen=outer,fillstyle=solid,fillcolor=myblue2](0,3.08)(16,2.08)
		\rput[Bl](2,2.34){\LARGE \textbf{Mean shift}}
		\rput[Bl](8,2.34){\LARGE 1685}	
		\rput[Bl](11,2.34){\LARGE 1.1 s}
		\rput[Bl](14,2.34){\LARGE \textbf{0.5 \%}}	
		%resultat SLIC
		\psframe[linewidth=0.04,dimen=outer,fillstyle=solid,fillcolor=myblue1](0,2.12)(16,1.12)
		\rput[Bl](2,1.38){\LARGE \textbf{SLIC}}
		\rput[Bl](8,1.38){\LARGE \textbf{1516}}	
		\rput[Bl](11,1.38){\LARGE 0.3 s}
		\rput[Bl](14,1.38){\LARGE 0.6 \%}	
		%resultat Veksler
		\psframe[linewidth=0.04,dimen=outer,fillstyle=solid,fillcolor=myblue2](0,1.16)(16,0.16)
		\rput[Bl](2,0.42){\LARGE \textbf{Veksler}}
		\rput[Bl](8,0.42){\LARGE 1536}	
		\rput[Bl](11,0.42){\LARGE 6.1 s}
		\rput[Bl](14,0.42){\LARGE 1.5 \%}	
	\end{pspicture}
	}
	\end{center}
\end{table}
Our comparison reveals that the two most interesting superpixel algorithms in our case are \emph{Felzenzswalb et al.}  \textit{et al.} method and \emph{SLIC} for their excecution time which are significantly shorter. Even if its execution time is slightly greater than the one of \emph{Felzenzswalb et al.} method, \emph{SLIC} produces significantly less superpixels making the next steps of SCIS faster. This low number of superpixels comes at the cost of an over-segmentation error of $0.6\%$ against $0.5\%$ with \emph{Felzenzswalb} method.

We chose  \emph{Felzenzswalb et al.} method because we think that an over-segmentation error, that the user cannot correct even if he adds a lot of seeds, is more annoying than waiting for processing a greater number of superpixels. 

\subsection{Superpixel features}
The choice of features describing superpixels is a crucial issue during the classification stage of SCIS. They must have two qualities:
\begin{itemize}
\item first, they have to ensure that any element of a wide variety of image could be distinguished from the rest of the picture;
\item second, they have to be fast to compute. 
\end{itemize}

As explained in section \ref{sec:previousWork}, some recent successful interactive segmentation methods \cite{santner2009interactive} use texture descriptors as textons or histograms of oriented gradients. Nevertheless, computing these kind of descriptors is still time and memory consuming. Moreover, increasing the number of features usually increases the required time for both learning and classification steps.  So, we simply describe each superpixel by its average color and the location of its center of mass. The experiments described in section \ref{sec:Eval} show that these features are quickly extracted while giving satisfactory results. 

We use RGB, the original color space of the images. We tested some other color spaces (Lab and HSI) but results were not significantly improved.

\subsection{Classifier}
Following the success of  \cite{santner2009interactive} and \cite{xu2008interactive}, we tested two classifiers: support vector machine (SVM) and random decision trees (RDT).

For SVM, we used C-SVM  libSVM implementation\footnote{\url{www.csie.ntu.edu.tw/~cjlin/libsvm/}} with a Radial Basis Function (RBF) kernel. With this kind of kernel, two parameters, the regularization parameter $C$ and the kernel parameter $\gamma$, must be tuned. 

For RDT, we used alglib implementation\footnote{\url{www.alglib.net/}}. Two parameters, the number of decision trees and the percentage of training data used to train each decision tree, must be given. 

\subsection{Selection of a  classifier}
\begin{figure*}[!t]
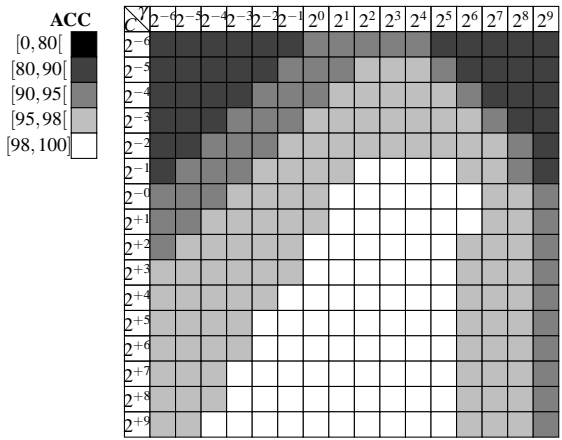
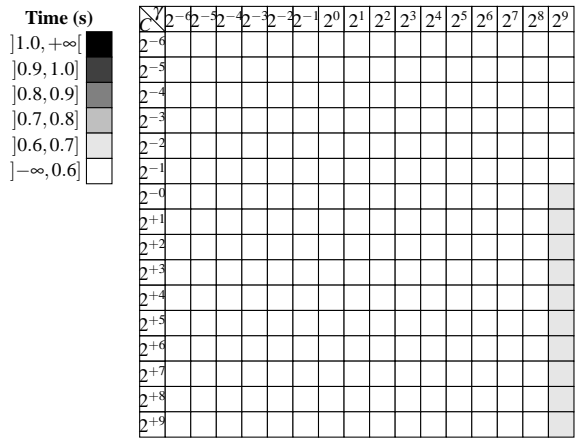
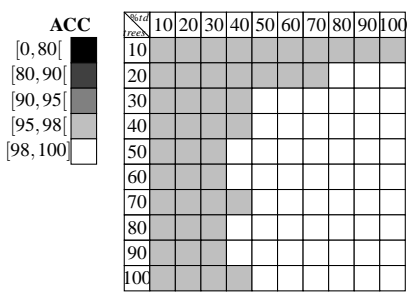
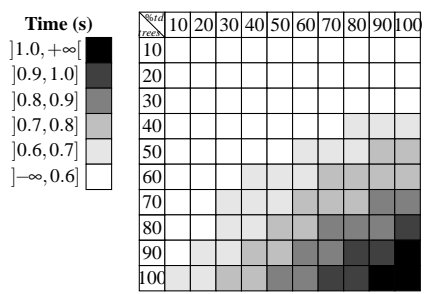

  %\begin{center}     
       \begin{subfigure}[t]{0.45 \textwidth}
       \begin{center}
		\scalebox{0.35} % Change this value to rescale the drawing.
		{
 
		}
       \end{center}
        \caption{\centering SVM: accuracy percentage.}
        \label{fig:SVMacc}
        \end{subfigure}
        \hfill       
       \begin{subfigure}[t]{0.45 \textwidth}
       
       \begin{center}
        \scalebox{0.35} % Change this value to rescale the drawing.
		{		
		%
 
		}
		\end{center}			
        \caption{\centering  SVM: execution time (in seconds).}
        \label{fig:SVMtime}
        \end{subfigure}   
        \\
        ~
        \\
        ~
        \\
       \begin{subfigure}[t]{0.45 \textwidth}
        
        	\begin{center}
			\scalebox{0.35} % Change this value to rescale the drawing.
			{
			%
 
		}		
       \end{center}
        \caption{\centering RDT: accuracy percentage.}
        \label{fig:RDTacc}
        \end{subfigure}
        \hfill
        \begin{subfigure}[t]{0.45 \textwidth}
        
       \begin{center}
		\scalebox{0.35} % Change this value to rescale the drawing.
		{		
		%
 
		}
       \end{center}
        \caption{\centering RDT: execution time (in seconds).}
        \label{fig:RDTtime}
        \end{subfigure}
        %\end{center}  
        \caption{Accuracy and execution time (in seconds) of two variants of SCIS using respectively SVM and RDT classifiers. For SVM variant we tried various values for $C$, the regularization parameter, and $\gamma$, the kernel parameter, searching for the pair giving the best results. For RDT variant we tried various values for $td$, the percentage of training data used to train each decision tree and $trees$, the number of decision trees, searching for the pair giving the best results.}      
     
        \label{fig:prPixels}
\end{figure*}

We obtained two variants of the method,namely: SCIS-SVM and SCIS-RDT and we evaluated their performances on a subset of Santner benchmark \cite{santner2010Interactive}. We selected randomly 25 images and, for each image, we chose some seeds, according to the two following rules:
\begin{itemize}
\item each semantic element of the image must have some seeds;
\item spatial distribution of seeds should be approximately uniform over the image. 
\end{itemize}

We tested each variant with different pairs of parameters, to analyze how its behavior evolves when parameters are modified. We used two criteria: the average execution time for an image and the accuracy (ACC) of the interactive method, regarding to the ground truth. Execution time is calculated thanks to the \textit{clock} function given in the C++ library \emph{time.h}, on a desktop PC featuring a 2.6 GHz Intel Core i7 processor. The accuracy is the proportion of true results among the total number of results:
\begin{equation}
ACC = \frac{ \displaystyle \sum_{1}^{N} T_{j} }{ \displaystyle  \sum_{1}^{N} T_{j} \cup F_{j}}
\end{equation}
with:
\begin{itemize}

\item $ T_{j} $ the number of true positives: pixels belonging to the class $j$ in both segmentation result and ground truth;
\item $ F_{j} $ the number of false positives: pixels belonging to the class $j$ in  segmentation result but not in the ground truth;
\item $ N $ the number of classes in the ground truth. 
\end{itemize}

Figures \ref{fig:RDTacc} and \ref{fig:RDTtime} show that, for RDT, accurate results are achieved with a  high number of decision trees and  using a substantial percentage of training data for each decision tree, which comes at the cost of a significantly increased execution time. The only case with both satisfactory accuracy and computational cost is when few decision trees are trained with a large part of training data. 

For SCIS-SVM variant,  according to Figures \ref{fig:SVMacc} and \ref{fig:SVMtime}, there are some pairs of values for parameters $\gamma$ and $C$ giving both a good accuracy and a fast classification, for example $\gamma=4$ and $C=4$, $\gamma=4$ and $C=8$, $\gamma=8$ and $C=8$, etc.

\begin{figure}
\begin{center}
\scalebox{0.5} % Change this value to rescale the drawing.
{
\begin{pspicture}(0,-3.6479166)(14.9383335,3.6879168)
\definecolor{color17324b}{rgb}{0.996078431372549,0.796078431372549,0.19607843137254902}
\definecolor{color17324}{rgb}{0.996078431372549,0.807843137254902,0.19607843137254902}
\definecolor{color8323b}{rgb}{0.07450980392156863,0.49411764705882355,0.8}
\definecolor{color8323}{rgb}{0.07450980392156863,0.4980392156862745,0.8}
\psframe[linewidth=0.04,linecolor=color17324,dimen=outer,fillstyle=solid,fillcolor=color17324b](12.145833,0.53791666)(11.745833,-2.9820833)
\psframe[linewidth=0.04,linecolor=color8323,dimen=outer,fillstyle=solid,fillcolor=color8323b](2.7458334,-1.9820833)(2.3458333,-2.9820833)
\psframe[linewidth=0.04,linecolor=color8323,dimen=outer,fillstyle=solid,fillcolor=color8323b](1.7458333,-1.7220833)(1.3458333,-2.9820833)
\psframe[linewidth=0.04,linecolor=color8323,dimen=outer,fillstyle=solid,fillcolor=color8323b](3.7458334,-0.56208336)(3.3458333,-2.9820833)
\psframe[linewidth=0.04,linecolor=color8323,dimen=outer,fillstyle=solid,fillcolor=color8323b](4.7458334,0.84791666)(4.3458333,-2.9820833)
\psframe[linewidth=0.04,linecolor=color8323,dimen=outer,fillstyle=solid,fillcolor=color8323b](5.7458334,-0.0020833348)(5.3458333,-2.9820833)
\psframe[linewidth=0.04,linecolor=color8323,dimen=outer,fillstyle=solid,fillcolor=color8323b](6.7458334,-1.1320833)(6.3458333,-2.9820833)
\psframe[linewidth=0.04,linecolor=color8323,dimen=outer,fillstyle=solid,fillcolor=color8323b](7.7458334,-1.4220834)(7.3458333,-2.9820833)
\psframe[linewidth=0.04,linecolor=color8323,dimen=outer,fillstyle=solid,fillcolor=color8323b](8.745833,0.68791664)(8.345834,-2.9820833)
\psframe[linewidth=0.04,linecolor=color8323,dimen=outer,fillstyle=solid,fillcolor=color8323b](9.745833,-1.6720834)(9.345834,-2.9820833)
\psframe[linewidth=0.04,linecolor=color8323,dimen=outer,fillstyle=solid,fillcolor=color8323b](10.745833,-1.8020834)(10.345834,-2.9820833)
\psframe[linewidth=0.04,linecolor=color17324,dimen=outer,fillstyle=solid,fillcolor=color17324b](13.145833,-0.78208333)(12.745833,-2.9820833)
\psline[linewidth=0.06cm](12.725833,-2.9620833)(13.525833,-2.9620833)
\psframe[linewidth=0.04,dimen=outer,fillstyle=solid,fillcolor=black](0.8458333,-2.8620834)(0.6458333,-3.0620832)
\psframe[linewidth=0.04,linecolor=color17324,dimen=outer,fillstyle=solid,fillcolor=color17324b](2.1458333,-1.2120833)(1.7458333,-2.9820833)
\psframe[linewidth=0.04,linecolor=color17324,dimen=outer,fillstyle=solid,fillcolor=color17324b](3.1458333,-2.2920833)(2.7458334,-2.9820833)
\psframe[linewidth=0.04,linecolor=color17324,dimen=outer,fillstyle=solid,fillcolor=color17324b](4.1458335,0.31791666)(3.7458334,-2.9820833)
\psframe[linewidth=0.04,linecolor=color17324,dimen=outer,fillstyle=solid,fillcolor=color17324b](5.1458335,2.8179166)(4.7458334,-2.9820833)
\psframe[linewidth=0.04,linecolor=color17324,dimen=outer,fillstyle=solid,fillcolor=color17324b](6.1458335,2.4179168)(5.7458334,-2.9820833)
\psframe[linewidth=0.04,linecolor=color17324,dimen=outer,fillstyle=solid,fillcolor=color17324b](7.1458335,-0.25208333)(6.7458334,-2.9820833)
\psframe[linewidth=0.04,linecolor=color8323,dimen=outer,fillstyle=solid,fillcolor=color8323b](11.745833,-0.08208334)(11.345834,-2.9820833)
\psframe[linewidth=0.04,linecolor=color8323,dimen=outer,fillstyle=solid,fillcolor=color8323b](12.745833,-1.1720834)(12.345834,-2.9820833)
\psframe[linewidth=0.04,linecolor=color17324,dimen=outer,fillstyle=solid,fillcolor=color17324b](11.145833,-1.3320833)(10.745833,-2.9820833)
\psframe[linewidth=0.04,linecolor=color17324,dimen=outer,fillstyle=solid,fillcolor=color17324b](8.145833,-1.1720834)(7.7458334,-2.9820833)
\psframe[linewidth=0.04,linecolor=color17324,dimen=outer,fillstyle=solid,fillcolor=color17324b](9.145833,3.0279167)(8.745833,-2.9820833)
\psframe[linewidth=0.04,linecolor=color17324,dimen=outer,fillstyle=solid,fillcolor=color17324b](10.145833,-1.3520833)(9.745833,-2.9820833)
\psline[linewidth=0.02cm,linestyle=dashed,dash=0.16cm 0.16cm](0.74583334,-1.9820833)(13.345834,-1.9820833)
\psline[linewidth=0.02cm,linestyle=dashed,dash=0.16cm 0.16cm](0.74583334,1.0179167)(13.345834,1.0179167)
\psline[linewidth=0.02cm,linestyle=dashed,dash=0.16cm 0.16cm](0.74583334,0.017916664)(13.345834,0.017916664)
\psline[linewidth=0.02cm,linestyle=dashed,dash=0.16cm 0.16cm](0.74583334,2.0179167)(13.345834,2.0179167)
\rput(0.74583334,-2.9620833){\psaxes[linewidth=0.06,ticks=none,ticksize=0.10583333cm,dx=1.0cm,dy=1.0cm,Dy=100,showorigin=false](0,0)(0,0)(12,6)}
\psline[linewidth=0.02cm,linestyle=dashed,dash=0.16cm 0.16cm](0.74583334,-0.9820833)(13.345834,-0.9820833)
\usefont{T1}{ptm}{m}{n}
\rput(14.125052,-2.8570833){Images}
\usefont{T1}{ptm}{m}{n}
\rput(0.8466146,3.5029166){time (s)}
\psframe[linewidth=0.04,linecolor=color8323,dimen=outer,fillstyle=solid,fillcolor=color8323b](13.945833,2.8179166)(13.545834,2.4179168)
\psframe[linewidth=0.04,linecolor=color17324,dimen=outer,fillstyle=solid,fillcolor=color17324b](13.945833,3.3979166)(13.545834,2.9979167)
\usefont{T1}{ptm}{m}{n}
\rput(14.498178,3.1829166){RDT}
\usefont{T1}{ptm}{m}{n}
\rput(14.510834,2.6229167){SVM}
\end{pspicture} 
}
\\
\scalebox{0.5} % Change this value to rescale the drawing.
{
\begin{pspicture}(0,-2.6579165)(15.673333,2.6979167)
\definecolor{color17324b}{rgb}{0.996078431372549,0.796078431372549,0.19607843137254902}
\definecolor{color17324}{rgb}{0.996078431372549,0.807843137254902,0.19607843137254902}
\definecolor{color8323b}{rgb}{0.07450980392156863,0.49411764705882355,0.8}
\definecolor{color8323}{rgb}{0.07450980392156863,0.4980392156862745,0.8}
\psframe[linewidth=0.04,linecolor=color17324,dimen=outer,fillstyle=solid,fillcolor=color17324b](12.145833,-1.4420834)(11.745833,-1.9920833)
\psframe[linewidth=0.04,linecolor=color8323,dimen=outer,fillstyle=solid,fillcolor=color8323b](2.7458334,1.5279167)(2.3458333,-1.9920833)
\psframe[linewidth=0.04,linecolor=color8323,dimen=outer,fillstyle=solid,fillcolor=color8323b](1.7458333,-1.0020833)(1.3458333,-1.9920833)
\psframe[linewidth=0.04,linecolor=color8323,dimen=outer,fillstyle=solid,fillcolor=color8323b](3.7458334,-1.6020833)(3.3458333,-1.9920833)
\psframe[linewidth=0.04,linecolor=color8323,dimen=outer,fillstyle=solid,fillcolor=color8323b](4.7458334,-1.4620833)(4.3458333,-1.9920833)
\psframe[linewidth=0.04,linecolor=color8323,dimen=outer,fillstyle=solid,fillcolor=color8323b](5.7458334,-0.76208335)(5.3458333,-1.9920833)
\psframe[linewidth=0.04,linecolor=color8323,dimen=outer,fillstyle=solid,fillcolor=color8323b](6.7458334,0.41791666)(6.3458333,-1.9920833)
\psframe[linewidth=0.04,linecolor=color8323,dimen=outer,fillstyle=solid,fillcolor=color8323b](7.7458334,1.1679167)(7.3458333,-1.9920833)
\psframe[linewidth=0.04,linecolor=color8323,dimen=outer,fillstyle=solid,fillcolor=color8323b](8.745833,-0.122083336)(8.345834,-1.9920833)
\psframe[linewidth=0.04,linecolor=color8323,dimen=outer,fillstyle=solid,fillcolor=color8323b](9.745833,-1.2420833)(9.345834,-1.9920833)
\psframe[linewidth=0.04,linecolor=color8323,dimen=outer,fillstyle=solid,fillcolor=color8323b](10.745833,0.53791666)(10.345834,-1.9920833)
\psframe[linewidth=0.04,linecolor=color17324,dimen=outer,fillstyle=solid,fillcolor=color17324b](13.145833,-1.1320833)(12.745833,-1.9920833)
\psframe[linewidth=0.04,dimen=outer,fillstyle=solid,fillcolor=black](0.8458333,-1.8720833)(0.6458333,-2.0720832)
\psframe[linewidth=0.04,linecolor=color17324,dimen=outer,fillstyle=solid,fillcolor=color17324b](2.1458333,-0.76208335)(1.7458333,-1.9920833)
\psframe[linewidth=0.04,linecolor=color17324,dimen=outer,fillstyle=solid,fillcolor=color17324b](3.1458333,2.2779167)(2.7458334,-1.9920833)
\psframe[linewidth=0.04,linecolor=color17324,dimen=outer,fillstyle=solid,fillcolor=color17324b](4.1458335,-1.1820834)(3.7458334,-1.9920833)
\psframe[linewidth=0.04,linecolor=color17324,dimen=outer,fillstyle=solid,fillcolor=color17324b](5.1458335,-0.9120833)(4.7458334,-1.9920833)
\psframe[linewidth=0.04,linecolor=color17324,dimen=outer,fillstyle=solid,fillcolor=color17324b](6.1458335,0.80791664)(5.7458334,-1.9920833)
\psframe[linewidth=0.04,linecolor=color17324,dimen=outer,fillstyle=solid,fillcolor=color17324b](7.1458335,-0.13208334)(6.7458334,-1.9920833)
\psframe[linewidth=0.04,linecolor=color8323,dimen=outer,fillstyle=solid,fillcolor=color8323b](11.745833,-1.5920833)(11.345834,-1.9920833)
\psframe[linewidth=0.04,linecolor=color8323,dimen=outer,fillstyle=solid,fillcolor=color8323b](12.745833,-1.5120833)(12.345834,-1.9920833)
\psframe[linewidth=0.04,linecolor=color17324,dimen=outer,fillstyle=solid,fillcolor=color17324b](11.145833,1.1679167)(10.745833,-1.9920833)
\psframe[linewidth=0.04,linecolor=color17324,dimen=outer,fillstyle=solid,fillcolor=color17324b](8.145833,1.5979167)(7.7458334,-1.9920833)
\psframe[linewidth=0.04,linecolor=color17324,dimen=outer,fillstyle=solid,fillcolor=color17324b](9.145833,-0.21208334)(8.745833,-1.9920833)
\psframe[linewidth=0.04,linecolor=color17324,dimen=outer,fillstyle=solid,fillcolor=color17324b](10.145833,-0.95208335)(9.745833,-1.9920833)
\psline[linewidth=0.02cm,linestyle=dashed,dash=0.16cm 0.16cm](0.74583334,-0.9920833)(13.345834,-0.9920833)
\psline[linewidth=0.02cm,linestyle=dashed,dash=0.16cm 0.16cm](0.74583334,1.0079167)(13.345834,1.0079167)
\psline[linewidth=0.02cm,linestyle=dashed,dash=0.16cm 0.16cm](0.74583334,0.007916665)(13.345834,0.007916665)
\usefont{T1}{ptm}{m}{n}
\rput(15.125052,-1.9670833){Images}
\usefont{T1}{ptm}{m}{n}
\rput(0.8466146,2.5129166){time (s)}
\psframe[linewidth=0.04,linecolor=color8323,dimen=outer,fillstyle=solid,fillcolor=color8323b](14.525833,0.98791665)(14.1258335,0.5879167)
\psframe[linewidth=0.04,linecolor=color17324,dimen=outer,fillstyle=solid,fillcolor=color17324b](14.525833,1.5679166)(14.1258335,1.1679167)
\usefont{T1}{ptm}{m}{n}
\rput(15.078177,1.3529167){RDT}
\usefont{T1}{ptm}{m}{n}
\rput(15.090834,0.79291666){SVM}
\psframe[linewidth=0.04,linecolor=color8323,dimen=outer,fillstyle=solid,fillcolor=color8323b](13.745833,-1.8520833)(13.345834,-1.9920833)
\rput(0.74583334,-1.9720833){\psaxes[linewidth=0.06,ticks=none,ticksize=0.10583333cm,dx=1.0cm,dy=1.0cm,Dy=100,Ox=12,showorigin=false](0,0)(0,0)(13,4)}
\psframe[linewidth=0.04,linecolor=color17324,dimen=outer,fillstyle=solid,fillcolor=color17324b](14.145833,-1.8020834)(13.745833,-1.9920833)
\psline[linewidth=0.06cm](13.665833,-1.9720833)(14.465834,-1.9720833)
\end{pspicture} 
}
\end{center}
\caption{Time (in seconds) taken by the user to segment each image, with each variant of SCIS. For each image, time for SCIS-SVM variant is given in blue and time for SCIS-RDT variant is given in yellow.}

\label{fig:UserTime}
\end{figure}
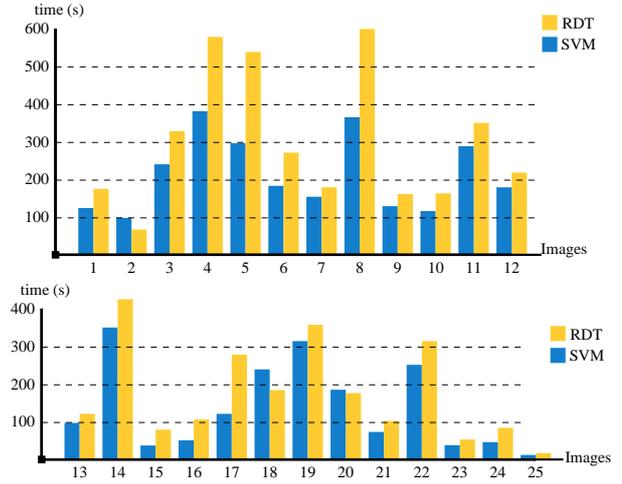

We used two more criteria to choose between the two classifiers: the average percentage of seeds and the average time taken by the user to obtain an accurate segmentation. The average percentage of seeds can be understood as the minimal percentage of pixels in an image which must be labeled by the user to obtain the final optimal segmentation. Of course it is impossible to be certain to have achieved this minimal percentage, so we simply tried to give as little seeds as possible, stopping when adding or removing seeds cannot improve the result. The average time taken by the user is the time required to give these minimal seeds or, in other words, the time between the first seed selection and the last seed selection, minus the execution time. We removed execution time in order to be independent of the classifier implementation. Our tests show that  SVM variant required at least $0.10\%$ of the pixels selected as seeds whereas RDT required at least $0.11\%$ of seeds. Moreover, Figure \ref{fig:UserTime} giving the user time for the 25 images shows that SVM  variant requires less user corrections to achieve segmentation. 

In conclusion, the variant of SCIS selected for the next experiments uses:
\begin{itemize}
\item \textit{Felzenzswalb et al.} algorithm to compute superpixels;
\item the average RGB color and the localization of its center of mass to describe each superpixel;
\item the SVM classifier with RBF kernel, $C=4, \gamma=4$.
\end{itemize}

This version of SCIS, implemented as a Gimp plug-in, is available here : \url{image.enfa.fr/scis}.
\section{Performance evaluation}
\label{sec:Eval}
\begin{sloppypar}
We compared SCIS to four interactive segmentation methods: the two most accurate methods in McGuinness review \cite{mcguinness2010comparative} (BPT and IGC), the Santner \textit{et al.} method \cite{santner2010Interactive} (TSRFTV)  and the Arbelaez method \cite{arbelaez2011contour} (CDHIS). All these algorithms are described in section \ref{sec:previousWork}. 
\end{sloppypar}

We use two benchmarks (images, ground truth and measures): McGuinness benchmark \cite{mcguinness2010comparative} and Santner benchmark \cite{santner2010Interactive}.  

\subsection{McGuiness benchmark}

\subsubsection{Experimental setup}
In \cite{mcguinness2010comparative}, McGuinness \textit{et al.} compare the efficiency of four interactive segmentation methods \cite{adams1994seeded,boykov2001interactive,friedland2005siox,salembier2000binary} to solve binary classification problems. The 96 images they use, are a subset of publicly available Berkeley Segmentation Dataset \cite{martin2001database}. Images have been chosen to be representative of a large variety of segmentation challenges. A set of 100 ground truth images have been created by humans, each one corresponding to an object (the foreground) extracted from the background. Binarization produced by interactive segmentation algorithms is evaluated with two measures: one for boundary accuracy, one for object accuracy. 

The boundary accuracy measure is given by:
\begin{equation}
ACC_{boundary}= 100 \dfrac{\sum_{x} \min (\tilde{B_{G}}(x),\tilde{B_{M}}(x) ) }{\sum_{x} \max (\tilde{B_{G}}(x),\tilde{B_{M}}(x) ) } 
\label{eq-accb}
\end{equation}
with $B_{G}$ and $B_{M}$ the internal border pixels for ground truth and algorithm segmentation result respectively, and $\tilde{B_{G}}$ and $\tilde{B_{M}}$ these same sets extended using fuzzy-set theory as described in \cite{mcguinness2010comparative}. 

The object accuracy measure is given by:
\begin{equation}
ACC_{region} = 100 \frac{| G_{O} \cap M_{O}|}{|G_{O} \cup M_{O}|}
\label{eq-accr}
\end{equation}
with $M_{O}$ the set of pixels labeled as object by the algorithm, $G_{O}$ the set of pixels labeled as object in the ground truth and $|S|$ the cardinality of set $S$. 
\begin{sloppypar}
We compared SCIS to the two best challengers of McGuinness \textit{et al.} review (IGC and BPT), as well as the Arbelaez \textit{et al.} method (CDHIS). 
We evaluated SCIS and CDHIS on the same images than McGuinness  and used the same measures and the same ground truth to evaluate SCIS results. Moreover, as McGuinness \textit{et al.}, we restricted the user to a maximum of two minutes per image. For a fair comparison, we implemented SCIS  as a Gimp plug-in, allowing the user to zoom and change brush diameter. Finally, to remove any ambiguities concerning the goal to achieve, we showed the ground truth to the user. The same seeds were given for SCIS and CDHIS method. \end{sloppypar}

\subsubsection{Results and discussion}

Table \ref{tab:resBenchMcGuinness} shows average object and boundary accuracy scores achieved by IGC \cite{boykov2001interactive}, BPT \cite{salembier2000binary}, CDHIS \cite{arbelaez2011contour} and SCIS. These scores correspond to the averages of object and boundary accuracies measured when the user has finished the segmentation or when the allocated time was up.  According to these two measures, SCIS outperform the other methods. 

\begin{table}
	\caption{Overall average boundaries accuracy and object accuracy (measures \ref{eq-accb} and \ref{eq-accr}). A high value of boundary accuracy indicates that boundary in the ground truth and in the segmentation results have the same shape. A high value of object accuracy indicates that pixels in the ground truth and in the segmentation results belong to the same classes.}
	\label{tab:resBenchMcGuinness}
	\begin{center}
	\scalebox{0.4} % Change this value to rescale the drawing.
	{
	\begin{pspicture}(0,0)(18,6)
		\newrgbcolor{myblue1}{0.68, 0.79, 0.86}
		\newrgbcolor{myblue2}{0.58, 0.69, 0.86}
		%premiere ligne
		\psframe[linewidth=0.04,dimen=outer,fillstyle=solid,fillcolor=lightgray](0,6)(18,4)		
		\rput[Bl](1,4.6){\Huge \textbf{Algorithm}}
		\rput[Bl](6,4.6){\Huge \textbf{$\mathbf{ACC_{boundary}}$}}	
		\rput[Bl](13,4.6){\Huge \textbf{$\mathbf{ACC_{object}}$}}								
		%resultats Felzenzswalb
		\psframe[linewidth=0.04,dimen=outer,fillstyle=solid,fillcolor=myblue1](0,4.04)(18,3.04)
		\rput[Bl](1,3.25){\Huge BPT}
		\rput[Bl](6,3.25){\Huge 78 \%}	
		\rput[Bl](13,3.25){\Huge  92 \%}
		%resultat meanshift
		\psframe[linewidth=0.04,dimen=outer,fillstyle=solid,fillcolor=myblue2](0,3.08)(18,2.08)
		\rput[Bl](1,2.25){\Huge IGC}
		\rput[Bl](6,2.25){\Huge 77 \% }	
		\rput[Bl](13,2.25){\Huge  92 \%}
		%resultat SLIC
		\psframe[linewidth=0.04,dimen=outer,fillstyle=solid,fillcolor=myblue1](0,2.12)(18,1.12)
		\rput[Bl](1,1.31){\Huge CDHIS}
		\rput[Bl](6,1.31){\Huge 70 \% }	
		\rput[Bl](13,1.31){\Huge  91 \%}
		%resultat Veksler
		\psframe[linewidth=0.04,dimen=outer,fillstyle=solid,fillcolor=myblue2](0,1.16)(18,0.16)
		\rput[Bl](1,0.35){\Huge \textbf{SCIS}}
		\rput[Bl](6,0.35){\Huge \textbf{82 \%}}	
		\rput[Bl](13,0.35){\Huge \textbf{94 \%}}
	\end{pspicture}
	}
\end{center}
\end{table}

\subsection{Santner benchmark}

\subsubsection{Experimental setup}
\label{sec-santnerexpsetup}
In \cite{santner2010Interactive}, Santner \textit{et al.} propose a new multiclass interactive segmentation method and give a benchmark to evaluate it. They use 243 images of $625 \times 391$ pixels, for which they created 262 ground truth segmentations.  So, for some image, several ground truth are provided, often with different numbers of regions. Santner \textit{et al.} chose the arithmetic average of the Dice evaluation score as  the performance criterion for their benchmark:
\begin{equation}
\label{eq:dice}
dice(I_{result},I_{groundtruth} ) =100 \sum_{i=1}^{N} 2\frac{|R_{i} \cap G_{i}|}{| R_{i} \cup G_{i}|}
\end{equation}
where $I_{result}$ is the result produced by the interactive segmentation method, $I_{groundtruth}$ the ground truth, $N$ the number of classes, $R_{i}$ the set of pixels of the $i^{\text{th}}$ class in the resulting segmentation and $G_{i}$ the set of pixels of the  $i^{\text{th}}$ class in the ground truth. 

Santner \textit{et al.} give seeds used for their evaluation. We also produce two other sets of seeds. The seeds of the first set, $Seeds_{SCIS}^{1}$, are produced following these two rules:
\begin{itemize}
\item each semantic element of the image must have some seeds;
\item spatial distribution of seeds should be approximately uniform over the image. 
\end{itemize}

The user is allowed to update seeds without time constraint, until the segmentation produced by SCIS become stable. These seeds correspond to the standard user interaction for interactive segmentation method.  Because their spatial distribution should be approximately uniform, they are more numerous than Santner seeds. 

The seeds of the second set, $Seeds_{SCIS}^{2}$, are produced  by following the same rules than seeds of $Seeds_{SCIS}^{1}$ set and by adding a new constraint: the user cannot draw strokes on the image, he can only select some isolated pixels and label them. This kind of user interaction is extremely tedious but seeds produced are significantly less numerous.

We compared Santner best results with SCIS and CDHIS methods. For both of them we computed average dice score with Santner seeds and with our seeds.

\subsubsection{Results and discussion}
\begin{table}
\caption{Overall dice for Santner benchmark (measure \ref{eq:dice}). A high dice value indicates that a lot of pixels belong to the same classes in the ground truth and in the segmentation method results. For CDHIS and SCIS we measure arithmetic average of the dice score with seeds provided by Santner \textit{et al.} and with our own seeds $Seeds_{SCIS}^{1}$ and $Seeds_{SCIS}^{2}$.}

\label{tab:resBenchSantner}
	\begin{center}
	\scalebox{0.35} % Change this value to rescale the drawing.
	{
	\begin{pspicture}(0,0)(23,5)
		\newrgbcolor{myblue1}{0.68, 0.79, 0.86}
		\newrgbcolor{myblue2}{0.58, 0.69, 0.86}
		%premiere ligne
		\psframe[linewidth=0.04,dimen=outer,fillstyle=solid,fillcolor=lightgray](0,5)(23,3)		
		\rput[Bl](1,3.8){\Huge \textbf{Algorithm}}
		\rput[Bl](6.5,3.8){\Huge \textbf{Santner seeds}	}
		\rput[Bl](13,3.8){\Huge $ \mathbf{Seeds_{SCIS}^{1}}$ }	
		\rput[Bl](18,3.8){\Huge $ \mathbf{Seeds_{SCIS}^{2}}$ }									
		%resultats TSRTV
		\psframe[linewidth=0.04,dimen=outer,fillstyle=solid,fillcolor=myblue1](0,3.08)(23,2.08)
		\rput[Bl](1,2.25){\Huge TSRFTV}
		\rput[Bl](6.5,2.25){\Huge \textbf{93 \%}}	
		\rput[Bl](13,2.25){\Huge \space \space-- \space }
		\rput[Bl](18,2.25){\Huge \space \space-- \space }
		%resultat CDHIS
		\psframe[linewidth=0.04,dimen=outer,fillstyle=solid,fillcolor=myblue2](0,2.12)(23,1.12)
		\rput[Bl](1,1.31){\Huge CDHIS}
		\rput[Bl](6.5,1.31){\Huge 91 \% }	
		\rput[Bl](13,1.31){\Huge  91 \%}
		\rput[Bl](18,1.31){\Huge  95 \%}
		%resultat SCIS
		\psframe[linewidth=0.04,dimen=outer,fillstyle=solid,fillcolor=myblue1](0,1.16)(23,0.16)
		\rput[Bl](1,0.35){\Huge \textbf{SCIS}}
		\rput[Bl](6.5,0.35){\Huge  82 \%}	
		\rput[Bl](13,0.35){\Huge \textbf{98 \%}}
		\rput[Bl](18,0.35){\Huge  \textbf{98 \%}}
	\end{pspicture}
	}
		\end{center}
\end{table}

\begin{figure*}[!t]
	  \captionsetup[subfigure]{oneside,margin={0.5cm,0.5cm}}
\begin{center}
	 \begin{subfigure}[t]{0.45\textwidth} 
	 \begin{center}
		\scalebox{0.3}{
			\includegraphics{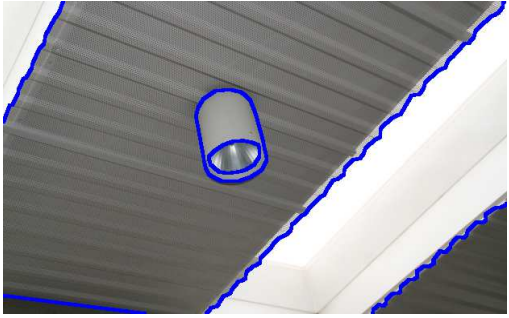}
		 }
	 \end{center}
        \caption{Image 0193 with region boundaries of ground truth 9060 of Santner benchmark. }
			\label{subfib:orgIm}
	\end{subfigure}
	~
    \begin{subfigure}[t]{0.45\textwidth} 
	 \begin{center}
		\scalebox{0.3}{
			\includegraphics{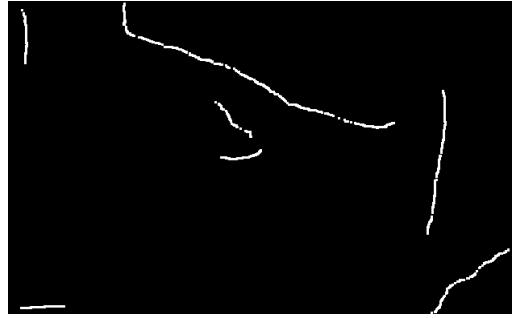}
		 }
		 \end{center}
        \caption{Santner seeds (775 pixels).}
			\label{subfib:santnerSeeds}
	\end{subfigure}
        \\ [2ex]  
        
        	\begin{subfigure}[t]{0.45\textwidth}
	 \begin{center}
		\scalebox{0.3}{
			\includegraphics{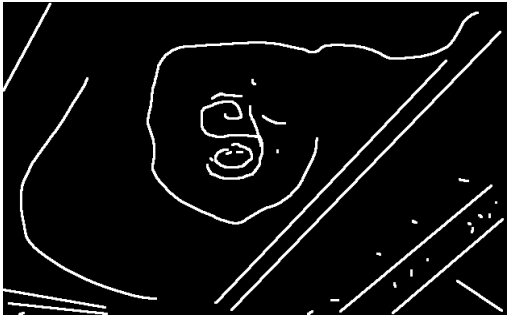}
        }
	 \end{center}
        \caption{Seeds (3175 pixels) of $Seeds_{SCIS}^{1}$ set.}
			\label{subfib:ourSeeds1}
        \end{subfigure}     
        ~
       \begin{subfigure}[t]{0.45\textwidth}
	 \begin{center}
		\scalebox{0.3}{
			\includegraphics{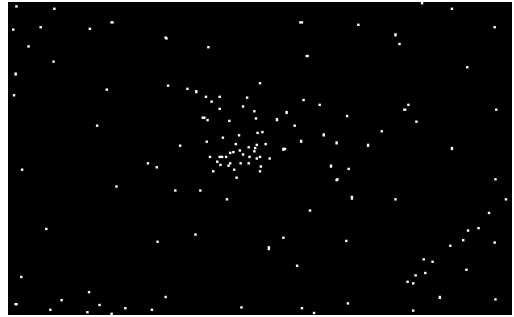}
        }
	 \end{center}
        \caption{Seeds (171 pixels) of $Seeds_{SCIS}^{2}$ set.}
			\label{subfib:ourSeeds2}
        \end{subfigure}
        
        \end{center}
        \caption{Differences between Santner seeds (\ref{subfib:santnerSeeds}) and our seeds (\ref{subfib:ourSeeds1} and  \ref{subfib:ourSeeds2}). Seeds are drawn in white. Santner seeds are less uniformly distributed over the image. To show a legible image we have increased seed size.}
        \label{fig:differencesSeeds}
\end{figure*}

Table \ref{tab:resBenchSantner} shows that seeds given by Santner, which are not uniformly distributed over the image, are not suitable for SCIS, for which location of pixel is a crucial information. Figure \ref{fig:differencesSeeds} shows an example where these differences are particularly apparent.

Seeds from $Seeds_{SCIS}^{1}$ or $Seeds_{SCIS}^{2}$ sets improve SCIS results. With these seeds, results produced by SCIS are more accurate than those produced by Santner method. Seeds of $Seeds_{SCIS}^{1}$ have two main differences with Santner seeds: their spatial distribution over the image is more uniform and they are more numerous. On the contrary, seeds of $Seeds_{SCIS}^{2}$ are less numerous than Santner ones ($0.09 \%$ of the image for our seeds and $0.37\%$ for Santner seeds) even if their spatial distribution is more uniform. Results computed with these seeds demonstrate that SCIS  dice score with $Seeds_{SCIS}^{1}$ is not due to the number of seeds but to their spatial distribution. So, in a context of a  comfortable interactive segmentation where short computation time is desirable, we think that SCIS is a good alternative to Santner and Arbelaez methods.

\subsection{Qualitative examples}
\begin{figure*}[!t]

	  \captionsetup[subfigure]{oneside,margin={6.5cm,0cm}}
\begin{center}
        
\begin{subfigure}[t]{\textwidth}

	 \begin{subfigure}[t]{0.95\textwidth}
		 \begin{center}
	 		\scalebox{0.3}{
				\includegraphics{DiversityOFUserIntentions_00}
			 }
	 	\end{center}
		\end{subfigure}
      \caption{Original image.}
\end{subfigure}
        \\[1ex]
          
\begin{subfigure}[t]{\textwidth}
	 \begin{subfigure}[t]{0.45\textwidth}
	 \begin{center}
		\scalebox{0.3}{
			\includegraphics{DiversityOFUserIntentions_01}
		 }
	 \end{center}
	\end{subfigure}
~
	\begin{subfigure}[t]{0.45\textwidth}
	 \begin{center}
		\scalebox{0.3}{
			\includegraphics{DiversityOFUserIntentions_02}
		}
	 \end{center}
	\end{subfigure}    
        \caption{Ground truth images.}
     \end{subfigure}
        \\[1ex]
          
\begin{subfigure}[t]{\textwidth}
	 \begin{subfigure}[t]{0.45\textwidth}
	 \begin{center}
		\scalebox{0.3}{
			\includegraphics{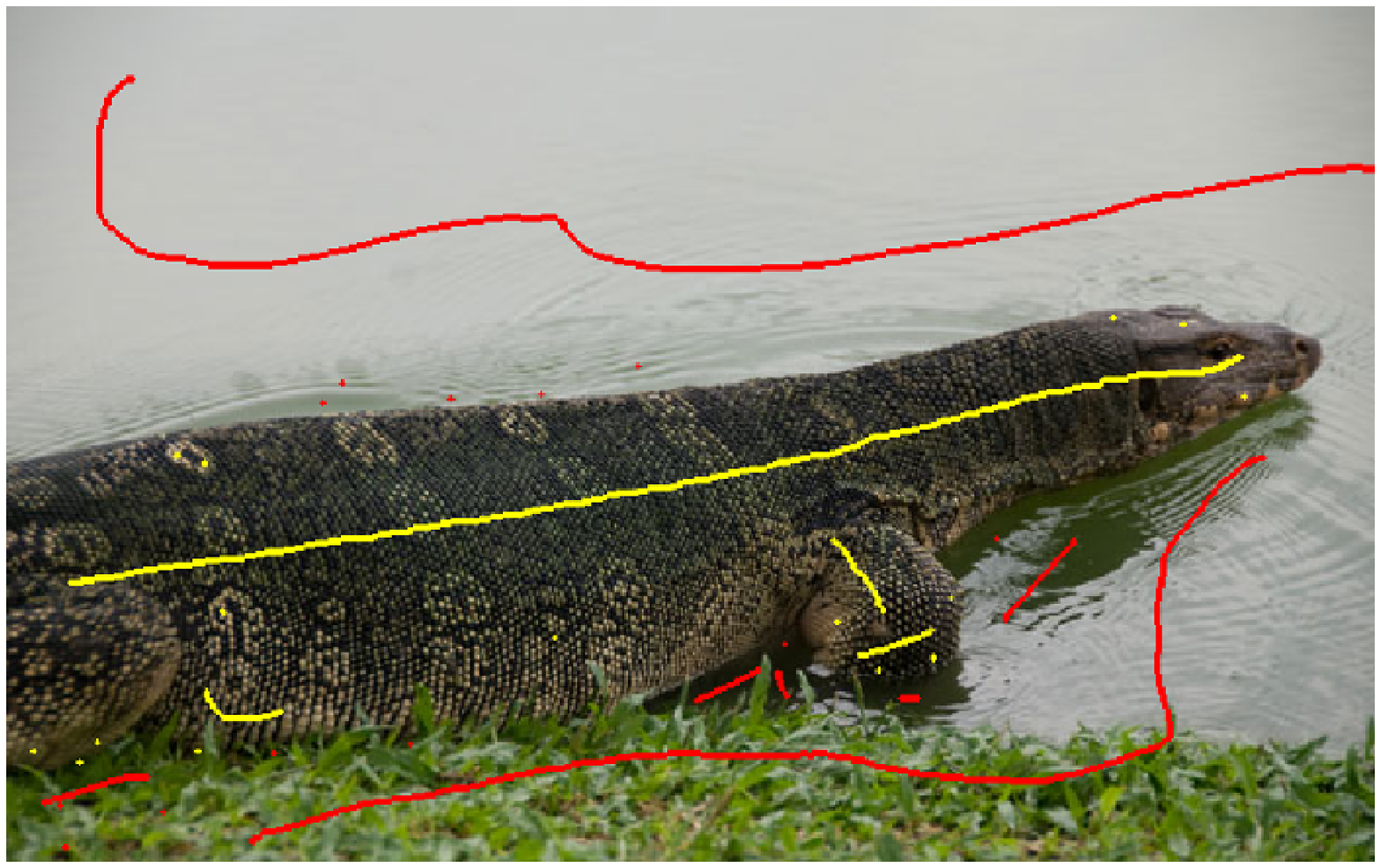}
		 }
	 \end{center}
	\end{subfigure}
~
	\begin{subfigure}[t]{0.45\textwidth}
	 \begin{center}
		\scalebox{0.3}{
			\includegraphics{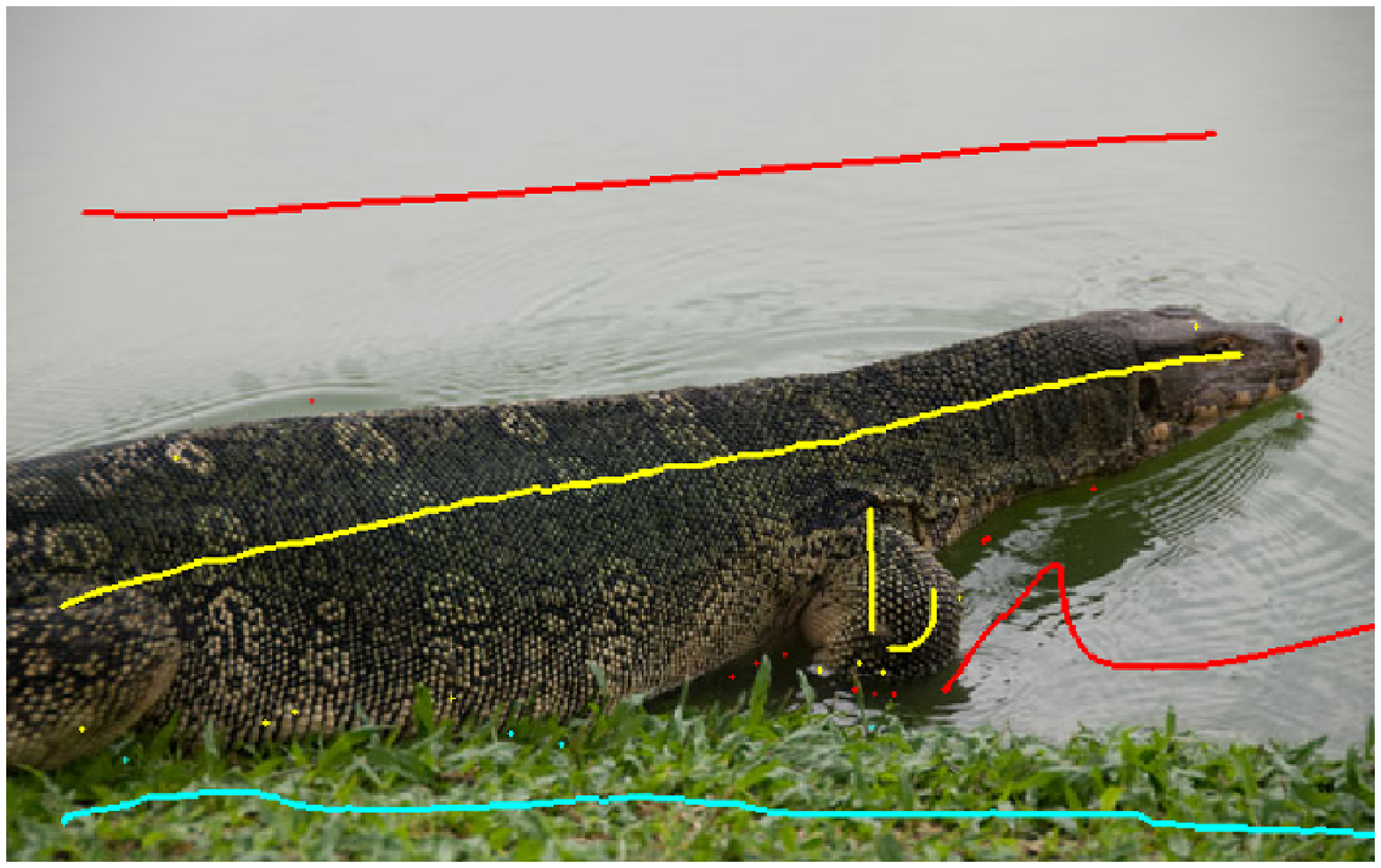}
		}
	 \end{center}
	\end{subfigure} 
        \caption{Seeds given by the user.}
     \end{subfigure}
        \\[1ex]
        
       \begin{subfigure}[t]{\textwidth}
	 \begin{subfigure}[t]{0.45\textwidth}
	 \begin{center}
		\scalebox{0.3}{
			\includegraphics{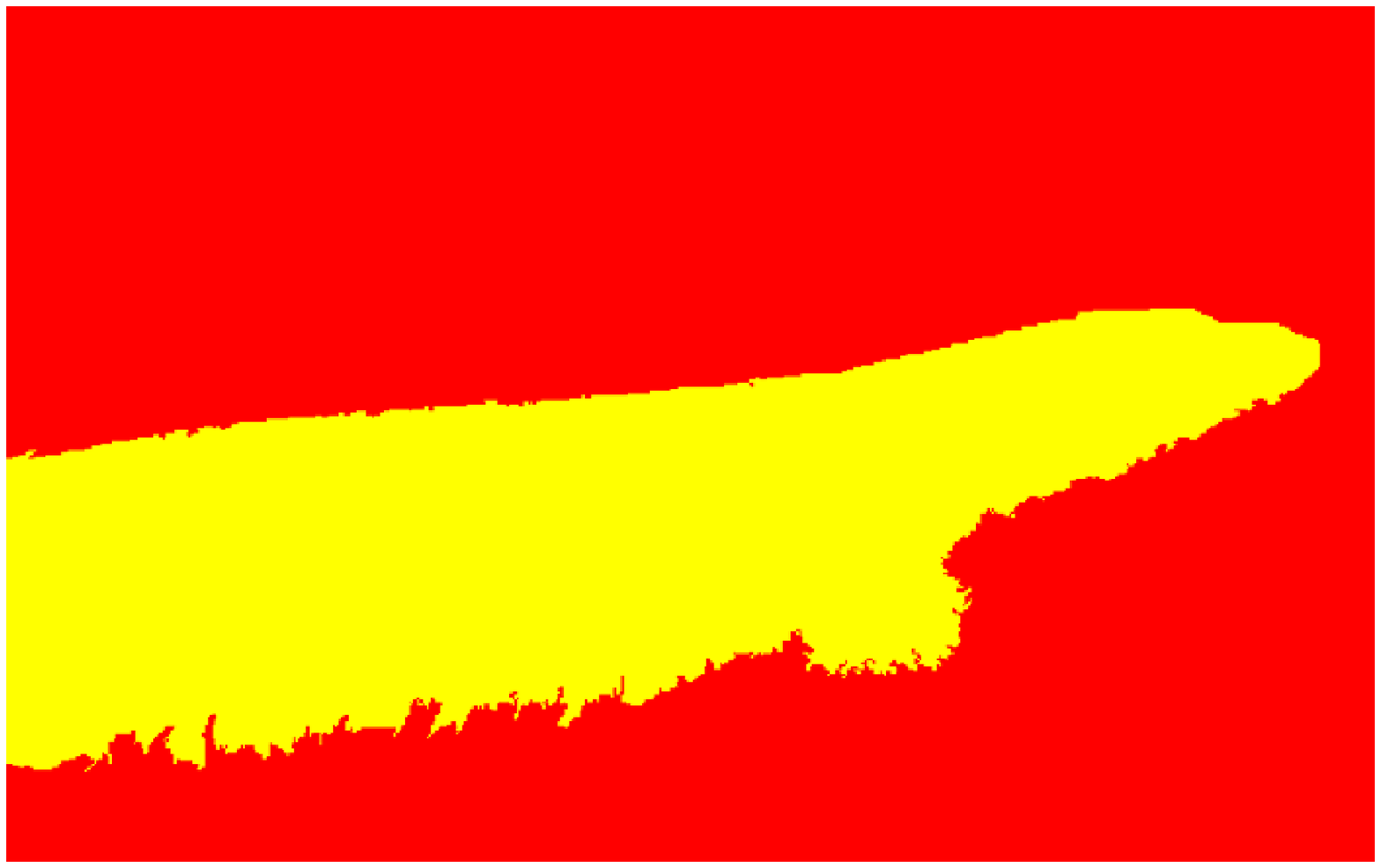}
		 }
	 \end{center}
	\end{subfigure}
~
	\begin{subfigure}[t]{0.45\textwidth}
	 \begin{center}
		\scalebox{0.3}{
			\includegraphics{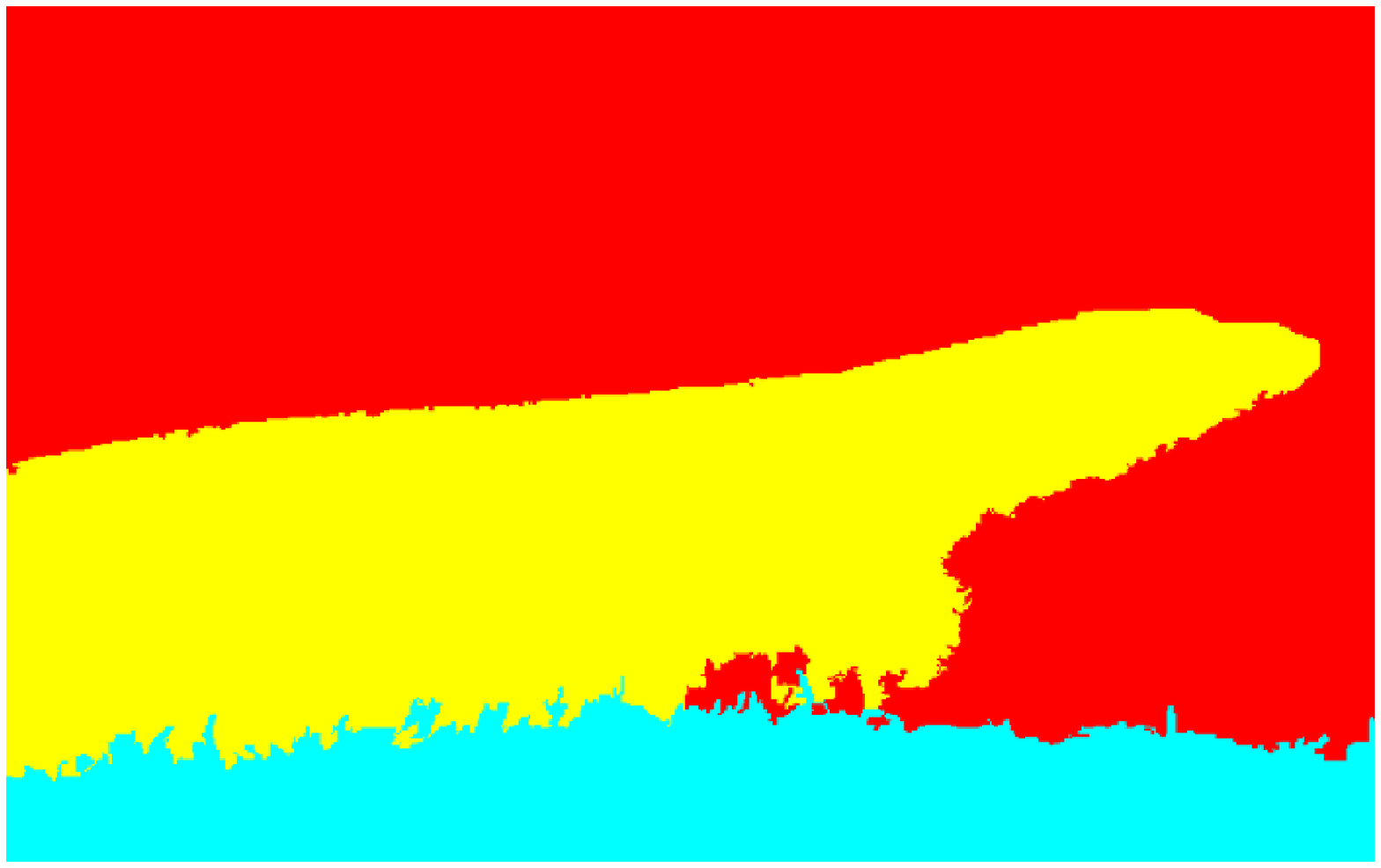}
		}
	 \end{center}
	\end{subfigure}
        \caption{Results obtained by SCIS.}
	\end{subfigure}
\end{center}
        \caption{ SCIS results for a same image, but with two different goals to achieve. The first column shows result achieved in a binarization context, whereas the second column shows result achieved in a semantic segmentation context. The first row shows the original image, the second row the segmentation to achieve, the third row the seeds given by the user and the fourth row the segmentation produced by SCIS.  }
        \label{fig:exemplesResultSCIS}
\end{figure*}

We conclude with an illustration of SCIS results for a same image, but in two different contexts of use: The first goal is a binarization, with the extraction of a main object (the animal) from the background, while the second goal is the segmentation of the image into semantic elements for the three following classes: grass, water and animal. Figure \ref{fig:exemplesResultSCIS} shows that for these two distinct goals, SCIS produces accurate segmentations, computing result for given seeds in less than one second.  

\section{Conclusion}
In this paper we proposed a new multiclass interactive segmentation method using superpixel classification. We tested various superpixel extraction algorithms and classifiers. By over-segmenting the image with \textit{Felzenzswalb and al.} algorithm, describing each superpixel by both its  average RGB color and its center of mass  and  by using a SVM classifier, we obtain an accurate and efficient interactive segmentation method\footnote{\url{image.enfa.fr/scis}}. We implemented this method as a Gimp plug-in and demonstrated its performance on both McGuinness and Santner benchmarks.

Nevertheless, even if the superpixel over-segmentation stage allows to quickly segment an image, errors in superpixels cannot be corrected in the final segmentation, regardless adding or removings seeds. This weakness of the algorithm could be frustrating for the user. So, in future work, we will try to overcome this limitation. A possibility to explore is computing a multi-level segmentation, for example by splitting superpixels with labels linked to several classes into several smaller superpixels.  

\subsection*{Acknowledgements}
We thank Pedro Felzenszwalb of Brown University, for allowing us to use his  implementation of its over-segmentation algorithm\footnote{\url{cs.brown.edu/~pff/segment/
}}. We also gratefully acknowledge Chih-Chung Chang and Chih-Jen Lin of National Taiwan University, for allowing us to use their excellent SVM training and classification library: libSVM \footnote{\url{www.csie.ntu.edu.tw/~cjlin/libsvm/}}. 

\bibliographystyle{plain} 

\bibliography{biblio2}
\end{document}